\documentclass[letterpaper]{article} 
\usepackage{aaai25}  
\usepackage{times}  
\usepackage{helvet}  
\usepackage{courier}  
\usepackage[hyphens]{url}  
\usepackage{graphicx} 
\usepackage{natbib}  
\usepackage{caption} 
\usepackage{array} 
\usepackage{makecell}
\usepackage{booktabs}
\usepackage{multirow}
\usepackage{amsmath}
\usepackage{amssymb}
\usepackage{pifont}
\usepackage{float}
\usepackage{subcaption}
\usepackage{graphicx}
\usepackage{multicol}
\usepackage{algorithmic}
\usepackage[ruled, vlined]{algorithm2e}

\usepackage{newfloat}
\usepackage{listings}
\DeclareCaptionStyle{ruled}{labelfont=normalfont,labelsep=colon,strut=off} 
\floatstyle{ruled}
\newfloat{listing}{tb}{lst}{}
\floatname{listing}{Listing}
\title{FedPop: Federated Population-based Hyperparameter Tuning}
\author {
    Haokun Chen\textsuperscript{\rm 1,}\textsuperscript{\rm 2}\footnote{Corresponding to haokun.chen@siemens.com and jindong.gu@outlook.com} \quad
    Denis Krompass\textsuperscript{\rm 2} \quad
    Jindong Gu\textsuperscript{\rm 3*} \quad
    Volker Tresp\textsuperscript{\rm 1,}\textsuperscript{\rm 4}
}
\affiliations {
    \textsuperscript{\rm 1} Ludwig Maximilian University of Munich \quad \textsuperscript{\rm 2} Siemens Technology \\
    \textsuperscript{\rm 3} University of Oxford \quad \textsuperscript{\rm 4} Munich Center for Machine Learning \\
}

\begin{document}
\maketitle
\begin{abstract}
Federated Learning (FL) is a distributed machine learning (ML) paradigm, in which multiple clients collaboratively train ML models without centralizing their local data. Similar to conventional ML pipelines, the client local optimization and server aggregation procedure in FL are sensitive to the hyperparameter (HP) selection. Despite extensive research on tuning HPs for centralized ML, these methods yield suboptimal results when employed in FL. This is mainly because their "training-after-tuning" framework is unsuitable for FL with limited client computation power. While some approaches have been proposed for HP-Tuning in FL, they are limited to the HPs for client local updates. In this work, we propose a novel HP-tuning algorithm, called Federated Population-based Hyperparameter Tuning (FedPop), to address this vital yet challenging problem. FedPop employs population-based evolutionary algorithms to optimize the HPs, which accommodates various HP types at both the client and server sides. Compared with prior tuning methods, FedPop employs an online  "tuning-while-training" framework, offering computational efficiency and enabling the exploration of a broader HP search space. Our empirical validation on the common FL benchmarks and complex real-world FL datasets, including full-sized Non-IID ImageNet-1K, demonstrates the effectiveness of the proposed method, which substantially outperforms the concurrent state-of-the-art HP-tuning methods in FL.
\end{abstract}

\section{Introduction}
Federated Learning (FL) is an effective machine learning paradigm suitable for decentralized data sources \cite{mcmahan2017communication}. Similar to the conventional ML algorithms, FL exhibits sensitivity to empirical choices of hyperparameters (HPs), such as learning rate, and optimization steps \cite{kairouz2021advances}. Hyperparameter Tuning (HPT) is a vital yet challenging component of the ML pipeline, which has been extensively studied in the context of centralized ML \cite{hutter2019automated}. However, traditional HPT methods, such as Bayesian Optimization \cite{snoek2012practical}, are not suitable for FL systems. These methods typically utilize the "training-after-tuning" framework. Within this framework, a substantial number of HPs needs to be evaluated, which involves repetitive training of models until convergence and subsequent retraining after optimizing the optimal HP. Such approaches can drastically increase the client's local computational costs and communication overheads, as it needs to execute multiple federated communications when evaluating only one HP. Furthermore, the distributed validation datasets impose a major challenge for HPT in FL, making it infeasible to evaluate HP for a large number of participating clients. 




Recently, a few approaches have emerged to address the problem intersection of HPT and FL, but they still exhibit certain limitations: FedEx \cite{khodak2021federated} pre-defines a narrower HP search space, while FLoRA \cite{zhou2023single} requires costly retraining after HP-optimization.  Moreover, they are only applicable for tuning the HPs used in client local updates. In this paper, we propose Federated Population-based Hyperparameter Tuning (\texttt{FedPop}) to address the challenge of tuning HPs for FL. \texttt{FedPop} applies population-based evolutionary algorithm \cite{jaderberg2017population} to optimize the HPs, which adds minimal computational overheads and accommodates various HP types at the client and server sides. Most importantly, \texttt{FedPop} employs an online "tuning-while-training" framework, enhancing efficiency and thereby allowing the exploration of a broader HP search space. 

In \texttt{FedPop}, we first construct multiple HP-configurations as our tuning population, i.e., we initialize multiple tuning processes (members) with randomly initialized HP-configuration, containing the HPs used in the server aggregation and the local client updates. Afterwards, we apply an evolutionary update mechanism to optimize the HPs of each member by leveraging information across different HP-configurations (\texttt{FedPop-G}). Hereby, the HPs in underperforming members will be replaced by a perturbed version of the HPs from better-performing ones, enabling an efficient and effective online propagation of the HPs. To further improve the HPs for the local client updates in a fine-grained manner, we consider the active clients in each communication round as our local population, where each member contains one HP-vector used in the local client update (\texttt{FedPop-L}). Similarly, evolutionary updates are executed based on the local validation performance of each member to tune these HP-vectors. Most importantly, all the tuning processes, i.e., members of the population, are decentralized and can be asynchronous, aligning perfectly with the distributed system design.

The proposed algorithm \texttt{FedPop} achieves new state-of-the-art (SOTA) results on three common FL benchmarks with both vision and language tasks, surpassing the concurrent SOTA HPT method for FL, i.e., FedEx \cite{khodak2021federated}. Moreover, we evaluate \texttt{FedPop} on large-scale cross-silo FL benchmarks with feature distribution shift \cite{li2021fedbn}, where its promising results demonstrate its applicability to complex real-world FL applications. Most importantly, we demonstrate the scalability of \texttt{FedPop}, where we show its applicability to full-sized ImageNet-1K \cite{deng2009imagenet} with ResNet-50 \cite{he2016deep}. Our contributions in this paper can be summarized as follows:
\begin{itemize}
    \item We propose an effective and efficient online hyperparameter tuning (HPT) algorithm, \texttt{FedPop}, to address HPT problem for decentralized ML systems. 
    \item We conduct comprehensive experiments on three common FL benchmarks with both vision and language tasks, in which \texttt{FedPop} achieves new SOTA results.
    \item We verify the maturity of \texttt{FedPop} for complex real-world cross-silo FL applications, and further analyze its convergence rate on full-sized non-IID ImageNet-1K, as well as its effectiveness when combined with various federated optimization algorithms.
\end{itemize}

\section{Related Works}
\label{sec:relatedwork}
\subsection{Hyperparameter Tuning for FL System} Previous works for tuning hyperparameters in FL focus only on specific aspects: \cite{wang2019adaptive} tunes only the local optimization epochs based on the client's resources, while \cite{koskela2018learning, mostafa2019robust, reddi2020adaptive} focus on the learning rate of client local training. \cite{dai2020federated, dai2021differentially} apply Bayesian Optimization (BO) \cite{snoek2012practical} in FL and optimize a personalized model for each client, while \cite{tarzanagh2022fednest} computes federated hypergradient and applies bilevel optimization. \cite{he2020towards, xu2020federated, garg2020direct, seng2022hanf, khan2023multi} tune architectural hyperparameters, in particular, adapt Neural Architecture Search (NAS) for FL. \cite{zhang2022fedtune} tunes hyperparameter based on the federated system overheads, while \cite{maumela2022population} assumes the training data of each client is globally accessible. \cite{mlodozeniec2023hyperparameter} partitions both clients and the neural network and tunes only the hyperparameters used in data augmentation. \cite{khodak2020weight, khodak2021federated} systematically analyze the challenges of hyperparameter tuning in FL and propose FedEx for client local hyperparameters. \cite{zhou2023single} proposes a hyperparameter optimization algorithm that aggregates the client's loss surfaces via single-shot upload. In contrast, the proposed method, FedPop, is applicable to various HP types on the client and server sides. In addition, it does not impose any restrictions on data volume and model architecture.

\subsection{Evolutionary Algorithms} Evolutionary algorithms are inspired by the principles of natural evolution, where stochastic genetic operators, e.g., mutation and selection, are applied to the members of the existing population to improve their survival ability, i.e., quality \cite{telikani2021evolutionary}. Evolutionary algorithms have shown their potential to improve machine learning algorithms, including architecture search \cite{real2017large, liu2017hierarchical}, hyperparameter tuning \cite{jaderberg2017population, parker2020provably}, and Automated Machine Learning (AutoML) \cite{liang2019evolutionary, real2020automl}. FedPop employs an online evolutionary algorithm, which is computationally efficient and explores a broader HP search space. To the best of our knowledge, FedPop is the first work combining evolutionary algorithms with HP optimization in Federated Learning. 

\section{Federated Hyperparmater Tuning}
\begin{figure}[t]
    \hspace{-7pt}
  \includegraphics[width=0.48\textwidth]{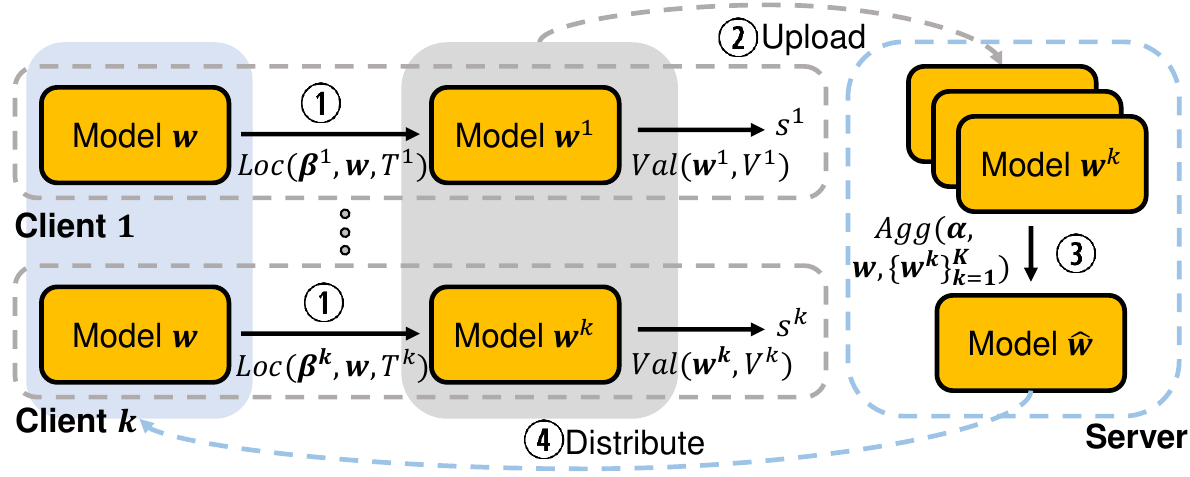}
  \centering
    \vspace{-17pt}
  \caption{Schematic illustration of the operations involved in one communication round, summarized as \texttt{Fed-Opt}.}
  \label{fig:fedopt}
    \vspace{-10pt}
\end{figure}

\subsection{Problem Definition}
In this section, we introduce the problem setup of hyperparameter tuning for FL. Following the setting introduced in \citep{khodak2021federated}, we assume that there are $N_c \in \mathbb{N}^{+}$ clients joining the federated communication. Each client $k$ owns a training, validation, and testing set, denoted by $T^k, V^k$, and $E^k$, respectively. To simulate the communication capacity of a real-world federated system, we presume that there are exactly $K \in \mathbb{N}^{+}$ active clients joining each communication round. In FedAvg \cite{mcmahan2017communication}, the central server obtains the model weight $\boldsymbol{w} \in \mathbb{R}^{d}$ by iteratively distributing $\boldsymbol{w}$ to the active clients and averaging the returned optimized weights, i.e., $\{\boldsymbol{w}^k | 1 \leq k \leq K\}$. 

More specifically, we denote the server aggregation and the client local training functions as \texttt{Agg} and \texttt{Loc}, respectively. Our goal is to tune the hyperparameter vectors (\textbf{HP-vectors}) used in these two functions. In particular, we denote the HP-vector used in \texttt{Agg} and \texttt{Loc} as $\boldsymbol{\alpha}$ and $\boldsymbol{\beta}$, which are sampled from the hyperparameter distribution $H_a$ and $H_b$, respectively. We define the combination of $\boldsymbol{\alpha}$ and $\boldsymbol{\beta}$ as one \textbf{HP-configuration}. In the following, we explain the general steps executed in the communication round, which involves these functions and HP-configurations. We summarize these steps as federated optimization (\texttt{Fed-Opt}), which is illustrated in Figure \ref{fig:fedopt}. Specifically, all active clients first execute function \texttt{Loc} (\ding{192}) in parallel:
\begin{equation}
\boldsymbol{w}^k \gets \texttt{Loc}(\boldsymbol{\beta}^k, \boldsymbol{w}, T^k),
\end{equation}
which takes the HP-vector $\boldsymbol{\beta}^k$, model parameters $\boldsymbol{w}$ distributed by the central server, and the local training set $T^k$ as inputs, and outputs the optimized model weight $\boldsymbol{w}^k$. Afterwards, the central server aggregates $\boldsymbol{w}^k$, uploaded by the active clients (\ding{193}) , and executes function \texttt{Agg} (\ding{194}):
\begin{equation}
\hat{\boldsymbol{w}} \gets \texttt{Agg}(\boldsymbol{\alpha}, \boldsymbol{w}, \{\boldsymbol{w}^k | 1 \leq k \leq K\}),
\end{equation}
which takes HP-vector $\boldsymbol{\alpha}$, current model parameter $\boldsymbol{w}$, updated model parameters from the active clients $\{\boldsymbol{w}^k | 1 \leq k \leq K\}$, and outputs the aggregated model weight $\hat{\boldsymbol{w}}$ which will be distributed to the active clients in the next communication round (\ding{195}). The goal of the federated hyperparameter tuning method is to find the optimal HP-vectors $\boldsymbol{\alpha}$ and $\boldsymbol{\beta}$ within a predefined communication budget.

\begin{figure}[t] 
  \includegraphics[width=0.47\textwidth]{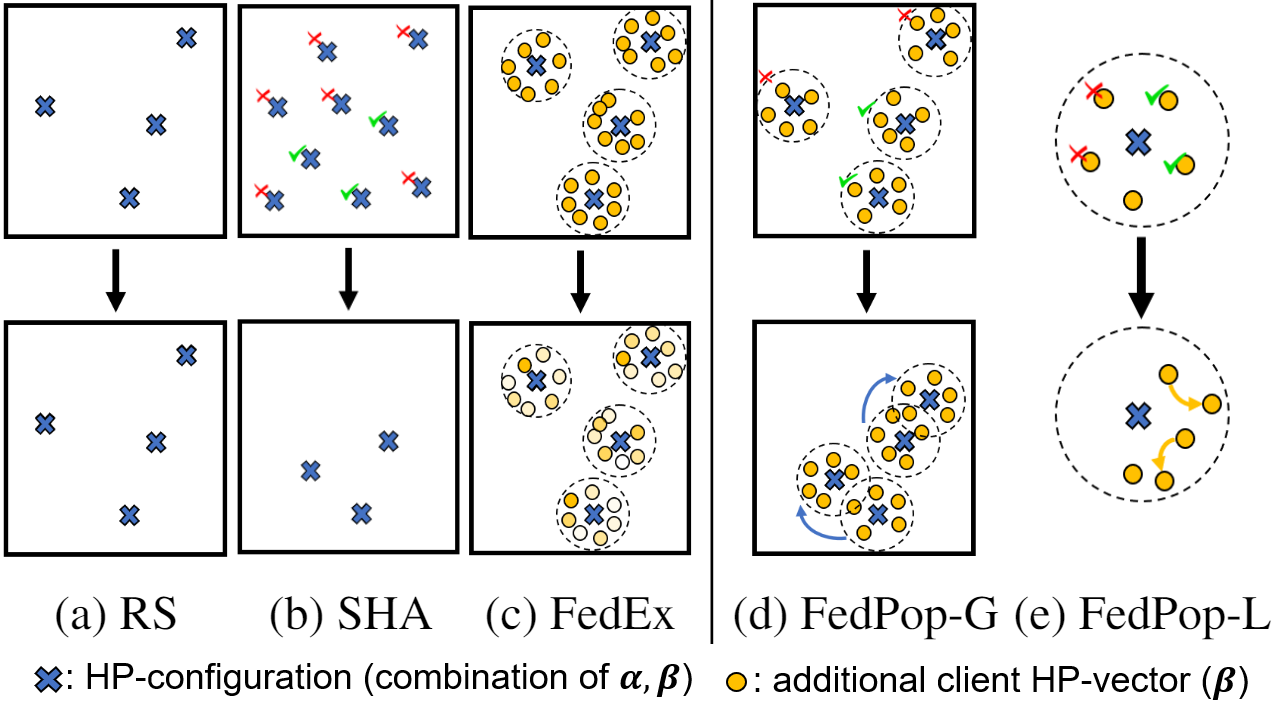}
    \vspace{-15pt}
  \captionof{figure}{Schematic comparison between \texttt{FedPop} and other baselines. \texttt{FedEx} optimizes the sampling probabilities of additional $\boldsymbol{\beta}$ based on validation performance (indicated by the brightness of the yellow dots). In contrast, our method supports the tuning of both server (\texttt{FedPop-G}) and clients (\texttt{FedPop-G} and \texttt{-L}) HP-vectors and explores broader search space with the help of evolutionary updates.}
  \label{fig:comparefig} 
    \vspace{-10pt}
\end{figure}

\subsection{Challenges in Federated Hyperparameter Tuning}
Given the problem defined in the previous section, we describe the two main challenges when tuning the hyperparameters for federated learning:

\textbf{(C1) Extrem resource limitations}: The communication budgets for optimizing ML models via FL are always very constrained due to the limited computational power of the clients and connection capacity of the overall system \cite{li2020federated}. Therefore, common hyperparameter tuning algorithms, such as extensive local hyperparameter tuning for each client, or experimenting multiple hyperparameter configurations for the overall federated system and then retraining, may not be suitable in the context of FL.

\textbf{(C2) Distributed validation data}: In centralized ML, most hyperparameter tuning algorithms select the HP-configurations based on their validation performance. However, the validation data ($V^k$) is distributed across the clients in FL. Computing a validation score over all clients is extremely costly and thus infeasible for FL. The alternative is to use the validation performance of client subsets, e.g., the active clients of the communication round, which greatly reduces computational costs. However, this may lead to evaluation bias when the client data are not independent and identically distributed (\emph{Non-IID}).

\subsection{Baseline Methods}

\begin{table}[t]
\caption{Number of HP-vectors tested in different HP-tuning methods on CIFAR-10 benchmark. \texttt{FedPop} experiments the largest number of HP-configurations among all methods. Detailed computations are provided in the Appendix.}
\vspace{-9pt}
\label{tab:numhps}
\setlength{\tabcolsep}{5mm}{
\begin{center}
\centering
\setlength\tabcolsep{5pt}
\begin{tabular}{c|cccc}
\toprule
Method & \makecell[c]{Number of \\ tried $\boldsymbol{\alpha}$} & \makecell[c]{Number of \\ tried $\boldsymbol{\beta}$} & \makecell[c]{Optim. \\ of $\boldsymbol{\alpha}$}  & \makecell[c]{Optim. \\ of $\boldsymbol{\beta}$} \\
\hline
\texttt{RS} & 5 & 5 & \ding{55} & \ding{55} \\
\texttt{SHA} & 27 & 27 & \ding{55} & \ding{55} \\
\texttt{FedEx} & 5 & 135 & \ding{55} & \ding{51} \\
\texttt{FedPop} & 45 & $>$1000 & \ding{51} & \ding{51}  \\
\bottomrule
\end{tabular}
\end{center}}
\vspace{-15pt}
\end{table}
Before introducing the proposed algorithm (\texttt{FedPop}) which addresses the challenges of HP-tuning in FL, we illustrate the adaptation of two widely adopted HP-tuning baselines for FL applications and their notations. For the FL setup, we define the maximum communication rounds for the FL system, i.e., total tuning budget, as $R_t$, and the number of initial HP-configurations as $N_c$, respectively. We devise two baseline methods for tuning $\boldsymbol{\alpha}, \boldsymbol{\beta}$ as follows:


(1) \textbf{Random Search (RS)} first initializes $N_c$ HP-configurations, resulting in a tuning budget of $R_c \ (= \frac{R_t}{N_c})$ for each HP-configuration. Afterwards, an ML model and $N_c$ tuning processes will be initialized, where each tuning process executes $R_c$ communication rounds to optimize the model using one specific HP-configuration. Finally, the optimized models from all tuning processes will be evaluated and the model exhibiting the highest testing accuracy, as well as its corresponding HP-configuration, are saved. 

(2) \textbf{Successive Halving (SHA)} is a variation of \texttt{RS} which eliminates $\frac{1}{\eta}$-quantile of the under-performing HP-configurations after specific numbers of communication rounds. Within the same tuning budget $R_t$, \texttt{SHA} is able to experiment more HP-configurations compared with \texttt{RS}, thus increasing the likelihood of achieving better results. Based on $R_t, N_c$, and the number of elimination operations, the time step for elimination can be computed. However, the elimination might also discard HP-configurations which lead to promising results but perform poorly at early stages.

\label{sec:proposmeh}
\begin{figure*}
  \centering
  \hspace{-10pt}
  \includegraphics[width=0.86\textwidth]{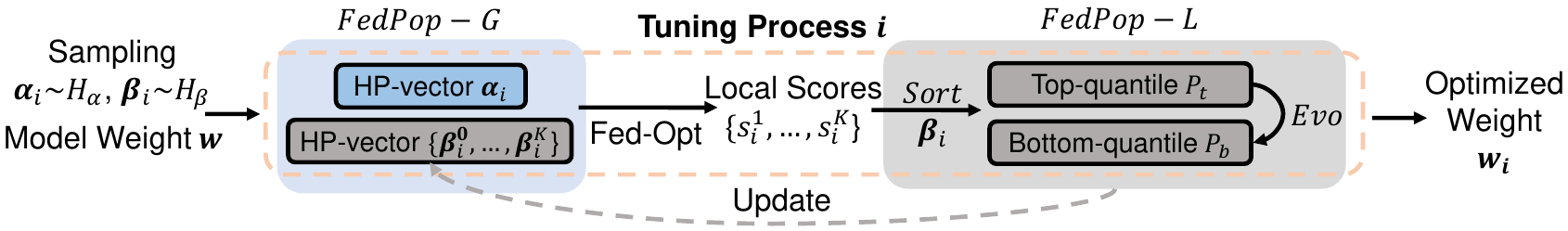}
    \makeatletter\def\@captype{figure}\makeatother\caption{Schematic illustration of \texttt{FedPop}, including \texttt{FedPop-L} for intra-configuration HP-tuning and \texttt{FedPop-G} for inter-configuration HP-tuning. \texttt{FedPop} employs an online "tuning-while-training" schema for tuning both server ($\boldsymbol{\alpha}$) and clients ($\boldsymbol{\beta}$) HP-vectors. All functions in \texttt{FedPop} can be executed in a parallel and asynchronous manner. \it{Best viewed in color.}} 
    \label{fig:main}
    \vspace{-12pt}
\end{figure*}

\textbf{Limitations:} These baseline methods exhibit two limitations when adapted to FL applications: First, as shown in Figure \ref{fig:comparefig}, their numbers of HP-configurations, as well as the HP values, are pre-defined and remain fixed throughout the tuning process. Second, these baseline methods are "static" and no active tuning is executed inside each tuning process. In other words, the model evaluation results are only obtained and utilized after a specific number of communication rounds. Therefore, we propose \texttt{FedPop}, a population-based tuning algorithm that updates the HP-configurations via evolutionary update algorithms. As a result of its high efficiency, it experiments the largest number of HP-vectors among all methods (Table \ref{tab:numhps}), which is introduced in the following.


\subsection{Proposed Method}

A schematic illustration of the proposed method, Federated Population-Based Hyperparameter Tuning (\texttt{FedPop}), is provided in Figure \ref{fig:main}. First, we randomly sample the HP-vectors ($\boldsymbol{\alpha}$ and $\boldsymbol{\beta}$) for each tuning process \emph{in parallel} and execute federated optimization \texttt{Fed-Opt} (Figure \ref{fig:fedopt}). Subsequently, we conduct \texttt{FedPop} based on the validation scores $s$ returned from the active clients in each tuning process. \texttt{FedPop} can be divided into 2 sub-procedures: \texttt{FedPop-G} aims at tuning both HP-vectors $\boldsymbol{\alpha}$ and $\boldsymbol{\beta}$ across all HP-configurations (\emph{inter-config}), while \texttt{FedPop-L} focuses on a fine-grained search of HP-vector $\boldsymbol{\beta}$ inside each HP-configurations (\emph{intra-config}).

\texttt{FedPop} can be wrapped with the aforementioned baselines. Specifically, the primary distinctions between the two wrappers are the number of initialized HP-configurations ($N_c$) and the execution of the tuning process eliminations in the intermediate steps. In the following, we use the most rudimentary method, \texttt{RS}, as our wrapper and elaborate on the proposed method. More details regarding \texttt{FedPop} wrapped with \texttt{SHA} are provided in the Appendix.  

With \texttt{RS} as the wrapper, \texttt{FedPop} first randomly initializes $N_c$ HP-configurations $(\boldsymbol{\alpha}_i, \boldsymbol{\beta}^0_i)$ as the initial population and copies the model weight vector $\boldsymbol{w}$. Afterwards, we randomly sample addition $K$ HP-vectors, i.e., $\{\boldsymbol{\beta}_i^k|1 \le k \le K\}$, inside a small $\Delta$-ball centered by $\boldsymbol{\beta}_i^0$. $\Delta$ is selected based on the distribution of the HP ($H_b$) and more details are provided in the Appendix. Directly sampling $\boldsymbol{\beta}_i^k$ from $H_b$ is problematic because we find that using too distinct HP-vectors for the active clients would lead to unstable model performance. This phenomenon was also observed by \cite{khodak2021federated}. We provide a schematic illustration of the sampling process in Figure \ref{fig:comparefig}, where the \emph{yellow dots} ($\{\boldsymbol{\beta}_i^k|1 \le k \le K\}$) are enforced to lie near the \emph{blue crosses} ($\boldsymbol{\beta}_i^0$). Note that this resampling process of $\boldsymbol{\beta}^k_i$ is also executed when $\boldsymbol{\beta}_i^0$ is perturbed via \texttt{Evo} in \texttt{FedPop-G}. Finally, $R_c$ communication rounds are executed for each tuning process in parallel, where the validation scores $s_i^k$, of the $k_{th}$ active client in the $i_{th}$ tuning process is recorded. The pseudo codes of the proposed method are given in Algorithm \ref{algo:fedpop}. 


\newcommand{\round}[1]{\ensuremath{\lfloor#1\rceil}}

\subsubsection{Evolution-based Hyperparameter Update (\texttt{Evo}):} Inspired by Population-based Training \cite{jaderberg2017population}, we design our evolution-based hyperparameter update function \texttt{Evo} as the following, 

\begin{equation}
\centering
\small
\texttt{Evo}(\boldsymbol{h}) = \left\{
\begin{aligned}
& \hat{h}_j \sim U(h_j - \delta_j, h_j + \delta_j) \quad \text{s.t.} \quad H_j = U(a_j, b_j),\\
& \hat{h}_j \sim U\{x^{i\pm\round{\delta_j}}_j, x^i_j\} \quad \text{s.t.} \left\{
    \begin{aligned}
    &H_j=U\{x^0_j, ..., x^n_j\}, \\
    &h_j = x_j^i, 
    \end{aligned}
    \right.
\end{aligned}
\right.    
\label{eq:evo}
\end{equation}

where $\boldsymbol{h}$ represents one HP-vector, i.e., $\boldsymbol{\alpha}$ or $\boldsymbol{\beta}$ for our problem setting. We perturb the $j_{th}$ value of $\boldsymbol{h}$, $h_j$, by resampling it from its possible neighboring values. Concretely, we select the new value of $h_j$ based on the type of its original sampling distribution $H_j$: (1) If $h_j$ is sampled from a continuous uniform distribution $H_j=U(a_j, b_j)$ (e.g., log-space of learning-rate, dropout), then we perturb $h_j$ by resampling it from $U(h_j - \delta_j, h_j + \delta_j)$, where $\delta_j \gets (b_j- a_j) \epsilon$ and $\epsilon$ is the pre-defined perturbation intensity. (2) If $h_j = x_j^i$ is sampled from a discrete uniform distribution $H_j=U\{x^0_j, ..., x^n_j\}$ (e.g., batch-size, epochs), then we perturb $h_j$ by reselecting its value from $\{x^{i-\round{\delta_j}}_j, x^i_j, x^{i+\round{\delta_j}}_j\}$. To further increase the diversity of the HP search space during tuning, we resample $h_j$ from its original distribution $H_j$ with probability $p_{re}$. 

While the HPs are randomly initialized in the early tuning stages, they become more informative as training progresses. To reflect this in \texttt{FedPop}, we employ a cosine annealing schema to control the values of $\epsilon$ and $p_{re}$ based on the conducted communication rounds $r$:

\begin{equation}
\centering
x_r = \frac{x_0}{2} \cdot (1 + cos(\pi \frac{r}{R_c})),
\end{equation}
where $x_r$ and $x_0$ denote the present and the initial value of the annealed parameter, respectively, $x$ is either $\epsilon$ or $p_{re}$.

\begin{figure}[b]
    \vspace{-8pt}
    \centering
  \includegraphics[width=0.475\textwidth]{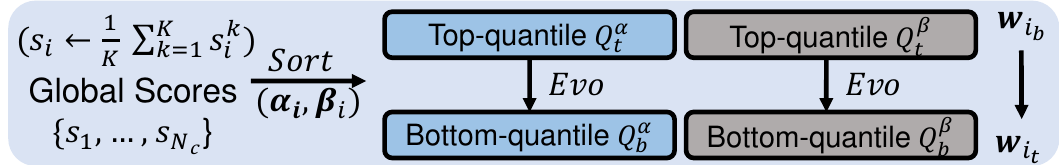}
    \vspace{-16pt}
  \caption{Schematic illustration of \texttt{FedPop-G}.}
  \vspace{-5pt}
  \label{fig:fedpopg}
\end{figure}

\begin{algorithm*}[t]
  \caption{Federated Population-Based Hyperparameter Tuning.}
    \label{algo:fedpop}
  \SetKwData{Left}{left}\SetKwData{This}{this}\SetKwData{Up}{up}
  \SetKwInOut{Input}{Input}\SetKwInOut{Output}{Output}
  \SetKwProg{Fn}{Function}{}{}
  \small
\textbf{Input:} Number of active clients per round $K$, number of HP-configurations $N_c$, total communication budget $R_t$, communication budget for each HP-configuration $R_c$ (computed by $\frac{R_t}{N_c}$), perturbation interval for \texttt{FedPop-G} $T_g$, initial model weight $\boldsymbol{w}$, $N_c$ server HP-vectors $\boldsymbol{\alpha} = \{\boldsymbol{\alpha}_1, ..., \boldsymbol{\alpha}_{N_c}\}$, $N_c$ client HP-vectors $\boldsymbol{\beta} = \{\boldsymbol{\beta}_1^0, ..., \boldsymbol{\beta}_{N_c}^0\}$.\\

Copy the model weights $\boldsymbol{w}_i \gets \boldsymbol{w}$ for all $N_c$ tuning processes.\\
\begin{multicols}{2}
\For{comm. round $\ r \leftarrow 1$ \KwTo $R_c$}{    
    \For {$ i \leftarrow 1$ \KwTo $N_c$ }{  
        \tcp{\textbf{in parallel}}
        \If {$len(\boldsymbol{\beta}_i)==1$}{
            Randomly sample $\{\boldsymbol{\beta}_i^k\}_{k=1}^K$ inside $\Delta$-ball of $\boldsymbol{\beta}_i^0$.
        }
        \For {Client $ k \leftarrow 1$ \KwTo $K$}{
            \tcp{\textbf{in parallel}}
            $\boldsymbol{w}^k_i \gets \texttt{Loc}(\boldsymbol{\beta}^k_i, \boldsymbol{w}_i, T^k)$\\
            $s_i^k \gets \texttt{Val}(\boldsymbol{w}_i^k, V^k)$
        }
        $\boldsymbol{\beta}_i \gets$ \texttt{FedPop-L} $(\boldsymbol{\beta}_i, \{s_i^k\}_{k=1}^K, K)$ \\
        $\boldsymbol{w}_i \gets \texttt{Agg}(\boldsymbol{\alpha}_i, \boldsymbol{w}_i, \{\boldsymbol{w}^k_i\})$ \\
        $s_i \gets \frac{1}{K}\sum_{k=1}^K s_i^k$ \\
        
    }
    \If {$r \% T_{g} = 0$}{
        $\{\boldsymbol{\alpha}_i, \boldsymbol{\beta}_i, \boldsymbol{w}_i\}_{i=1}^{N_c} \gets$ \texttt{FedPop-G} $(\{\boldsymbol{\alpha}_i, \boldsymbol{\beta}_i, \boldsymbol{w}_i, s_i\}_{i=1}^{N_c}, N_c)$ 
    }
}
\Return $\{\boldsymbol{w}_i\}_{i=1}^{N_c}$\\

~\\
\Fn{FedPop-L$(\boldsymbol{\beta}, \boldsymbol{s}, K)$}
{
    $\boldsymbol{P}_{b} \gets \{ k: s^k \geq \frac{\rho-1}{\rho}$-quantile$(\{s^k\})\}$\\
    $\boldsymbol{P}_{t} \gets \{ k: s^k \leq \frac{1}{\rho}$-quantile$(\{s^k\})\}$\\
    \For{$k_b \in \boldsymbol{P}_{b}$}{
        Sample $k_t$ from $\boldsymbol{P}_{t}$. \\
        Delete $\boldsymbol{\beta}^{k_b}$. \\
        $\boldsymbol{\beta}^{k_b} \gets \texttt{Evo}(\boldsymbol{\beta}^{k_t})$ \\   
    }
    \Return $\boldsymbol{\beta}$\\
}

\Fn{FedPop-G$(\boldsymbol{\alpha}, \boldsymbol{\beta}, \boldsymbol{w}, \boldsymbol{s}, N_c)$}
{
    $\boldsymbol{Q}_{b} \gets \{ i: s_i \geq \frac{\rho-1}{\rho}$-quantile$(\{s_i\}\}$\\
    $\boldsymbol{Q}_{t} \gets \{ i: s_i \leq \frac{1}{\rho}$-quantile$(\{s_i\})\}$\\
    \For{$i_b \in \boldsymbol{Q}_{b}$}{
        Sample $i_t$ from $\boldsymbol{Q}_{t}$. \\
        Delete $\boldsymbol{\alpha}_{i_b}, \boldsymbol{\beta}_{i_b}, \boldsymbol{w}_{i_b}$. \\
        $\boldsymbol{\alpha}_{i_b}, \boldsymbol{\beta}_{i_b}^0 \gets \texttt{Evo}(\boldsymbol{\alpha}_{i_t}, \boldsymbol{\beta}_{i_t}^0)$ \\
        $\boldsymbol{w}_{i_b} \gets \boldsymbol{w}_{i_t}$   \\
    }
    \Return $\boldsymbol{\alpha}, \boldsymbol{\beta}, \boldsymbol{w}$\\
}
\end{multicols}
\vspace{0.5em}
\end{algorithm*}

\subsubsection{\texttt{FedPop-G} for Inter-configuration Tuning:}
In \texttt{FedPop-G}, we adopt the average validation loss of all active clients, i.e., $s_i = \frac{1}{K} \sum_{k=1}^K s_i^k$, as the performance score for $i_{th}$ HP-configuration. However, $s_i$ may be a biased performance measurement, i.e., the disparity in the difficulty of the validation sets between different clients may lead to noisy $s_i$. To reduce the impact of the noise, \texttt{FedPop-G} is conducted with an interval of $T_g$ communication rounds. Hereby, the list of scores $s_i$ over $T_g$ rounds are recorded and their weighted sum with a power-law weight decay ($\gamma_g$) is utilized as the final measurement:
\vspace{-3pt}
\begin{equation}
\centering
s_i = \frac{\sum_{r=1}^{T_g} \gamma_g^{T_g - r} \cdot s_i^{(r)}}{\sum_{r=1}^{T_g} \gamma_g^{T_g - r}}.
\end{equation}

The tuning procedure starts by sorting the HP-configurations according to their validation scores. Afterwards, 2 subsets, i.e., $\boldsymbol{Q}_b$ and $\boldsymbol{Q}_t$, are constructed, representing the indices of the bottom and top $\frac{1}{\rho}$-quantile of the HP-configurations, respectively. Finally, the HP-configurations with indices in $\boldsymbol{Q}_b$ will be replaced by the perturbed version of the HP-configurations with indices in $\boldsymbol{Q}_t$. Specifically, $\boldsymbol{\alpha}_{i_b}, \boldsymbol{\beta}_{i_b}^0$ are replaced by the perturbed version of $\boldsymbol{\alpha}_{i_t}, \boldsymbol{\beta}_{i_t}^0$ via \texttt{Evo} (Equation \ref{eq:evo}), the model weight in $i_b$-$th$ HP-configuration ($\boldsymbol{w}_{i_b}$) are replaced by the $i_t$-$th$ ($\boldsymbol{w}_{i_t}$).
 
\begin{table*}[t]
\caption{Evaluation results of different hyperparameter tuning algorithms on three benchmark datasets. We report the \emph{global} and locally \emph{finetuned} (in the brackets) model performance with mean {\tiny±std} from 5-trial runs. The best results are marked in \textbf{bold}.}
\vspace{-10pt}
\label{tab:results}
\setlength{\tabcolsep}{1mm}{
\begin{center}
\scriptsize
\centering
\setlength\tabcolsep{10pt}
\renewcommand\arraystretch{1.01}
\begin{tabular}{c|c|ccc|cc|cc}

\toprule
\multirow{2}{*}{\makecell[c]{Tuning \\ Wrapper}} & \multirow{2}{*}{\makecell[c]{Tuning \\ Algorithm}} & \multicolumn{3}{c|}{CIFAR-10} & \multicolumn{2}{c|}{FEMNIST}                      &    \multicolumn{2}{c}{Shakespeare}              \\
\cline{3-9}
~ & ~ & IID & NIID ($Dir_{1.0}$)  & NIID ($Dir_{0.5}$) & IID & NIID & IID & NIID  \\

\hline
\multirow{5}{*}{RS}   & None & \makecell[c]{69.04 {\tiny ±7.38}\\(65.28 {\tiny ±5.83})} & \makecell[c]{63.47 {\tiny ±3.14} \\ (60.51 {\tiny ±8.03}) } & \makecell[c]{62.88 {\tiny ±8.13}\\(60.65 {\tiny ±7.37})} & \makecell[c]{82.86 {\tiny ±1.24}\\(83.76 {\tiny ±3.56})} & \makecell[c]{79.06 {\tiny ±5.59} \\ (83.09 {\tiny ±2.64}) } & \makecell[c]{33.76 {\tiny ±11.27}\\(31.19 {\tiny ±10.18})} & \makecell[c]{32.67 {\tiny ±12.27}\\(31.32 {\tiny ±9.92})}\\

\cline{2-9}
~ & FedEx & \makecell[c]{67.91 {\tiny ±7.15}\\(64.21 {\tiny ±7.84})}  & \makecell[c]{64.34 {\tiny ±5.28}\\(62.97 {\tiny ±7.27})} & \makecell[c]{63.22 {\tiny ±7.13}\\(61.92 {\tiny ±8.06})} & \makecell[c]{82.84 {\tiny ±0.80}\\(82.57 {\tiny ±3.25})} & \makecell[c]{82.14 {\tiny ±1.60}\\(84.03 {\tiny ±2.48})} & \makecell[c]{42.68 {\tiny ±7.24}\\(41.22 {\tiny ±6.34})} & \makecell[c]{44.28 {\tiny ±8.78}\\(46.69 {\tiny ±7.39})} \\

\cline{2-9}
~ & \textbf{FedPop} & \makecell[c]{\textbf{71.18} {\tiny ±4.68}\\(\textbf{68.01} {\tiny ±3.42})} & \makecell[c]{\textbf{68.25} {\tiny ±5.03}\\(\textbf{65.74} {\tiny ±3.97})} & \makecell[c]{\textbf{67.01} {\tiny ±4.98}\\(\textbf{65.24} {\tiny ±3.97})}  & \makecell[c]{\textbf{84.33} {\tiny ±1.41}\\(\textbf{85.99} {\tiny ±1.62})} & \makecell[c]{\textbf{83.21} {\tiny ±2.08}\\(\textbf{85.48} {\tiny ±1.48})} & \makecell[c]{\textbf{44.30} {\tiny ±3.37}\\(\textbf{44.46} {\tiny ±3.53})} & \makecell[c]{\textbf{47.28} {\tiny ±3.47}\\(\textbf{50.25} {\tiny ±3.87})}\\


\hline
\hline
\multirow{5}{*}{SHA}   & None & \makecell[c]{78.57 {\tiny ±2.39} \\ (75.93 {\tiny ±4.96}) } & \makecell[c]{70.37 {\tiny ±5.03} \\ (67.83 {\tiny ±4.41})}& \makecell[c]{68.65 {\tiny ±4.68} \\ (65.58 {\tiny ±8.10}) }  & \makecell[c]{83.81 {\tiny ±0.45}\\(85.52 {\tiny ±1.63})} & \makecell[c]{80.62 {\tiny ±2.88}\\(87.64 {\tiny ±0.64})}  & \makecell[c]{52.23 {\tiny ±2.54}\\(49.06 {\tiny ±5.98})} & \makecell[c]{51.68 {\tiny ±0.95}\\(48.83 {\tiny ±3.12})} \\

\cline{2-9}
~ & FedEx   & \makecell[c]{79.83 {\tiny ±2.59} \\ (77.04 {\tiny ±1.45}) } & \makecell[c]{72.02 {\tiny ±4.91} \\ (70.81 {\tiny ±4.65}) } & \makecell[c]{69.69 {\tiny ±7.03} \\ (67.02 {\tiny ±7.65}) }  & \makecell[c]{81.19 {\tiny ±3.24}\\(85.69 {\tiny ±1.91})} & \makecell[c]{82.76 {\tiny ±0.54}\\(86.79 {\tiny ±2.89})}  & \makecell[c]{51.79 {\tiny ±1.25}\\(51.89 {\tiny ±1.30})} & \makecell[c]{51.26 {\tiny ±2.73}\\(51.01 {\tiny ±3.36})}  \\

\cline{2-9}
~ & \textbf{FedPop}   & \makecell[c]{\textbf{81.47} {\tiny ±1.24} \\ (\textbf{78.96} {\tiny ±0.87})} & \makecell[c]{\textbf{76.42} {\tiny ±3.04} \\ (\textbf{75.03} {\tiny ±2.56}) } &  \makecell[c]{\textbf{74.88} {\tiny ±2.06} \\ (\textbf{72.41} {\tiny ±1.87}) }   & \makecell[c]{\textbf{84.33} {\tiny ±0.57} \\ (\textbf{86.84} {\tiny ±0.98})} & \makecell[c]{\textbf{83.26} {\tiny ±0.86} \\ (\textbf{88.33} {\tiny ±0.79})} & \makecell[c]{\textbf{53.48} {\tiny ±0.57} \\ (\textbf{52.66} {\tiny ±1.91}) } & \makecell[c]{\textbf{53.07} {\tiny ±0.97} \\ (\textbf{52.79} {\tiny ±0.36}) } \\


\bottomrule       
\end{tabular}
\end{center}}
\vspace{-13pt}
\end{table*}
\subsubsection{\texttt{FedPop-L} for Intra-configuration Tuning:}
To further explore the local neighborhood of $\boldsymbol{\beta}_i^0$ for client local update in a fine-grained manner, we apply \texttt{FedPop-L} inside each tuning process. Hereby, we provide an informative assessment of $\boldsymbol{\beta}_i^0$ and its local neighborhood to enhance the robustness of HP-configuration. For simplicity, we omit $i$ in the following notations. We consider the base HP-vector $\boldsymbol{\beta}^0$ as the perturbation center and restrict the perturbated HP-vector to lie inside a $\Delta$-ball of it, i.e., $||\boldsymbol{\beta}^k - \boldsymbol{\beta}^0||_2 \leq \Delta$. At each communication round, $\boldsymbol{\beta}^k$ will be assigned to \texttt{Loc} of the $k_{th}$ active client, the validation loss of the optimized model $\boldsymbol{w}^k$ will be recorded as the score $s^k$ for HP-vector $\boldsymbol{\beta}^k$. Afterwards, $\{\boldsymbol{\beta}^k\}_{k=1}^K$ will be sorted according to the validation scores and separated into 2 subsets, containing the indices of the bottom ($\boldsymbol{P}_b$) and the top ($\boldsymbol{P}_t$) $\frac{1}{\rho}$-quantile of the $\boldsymbol{\beta}$, respectively. Finally, the HP-vectors $\boldsymbol{\beta}^{k_t}$ with indices in $\boldsymbol{P}_t$ will be perturbed to replace the HP-vectors $\boldsymbol{\beta}^{k_b}$ with indices in $\boldsymbol{P}_b$ via \texttt{Evo}. 

\subsubsection{Solutions to Challenges:}
(\textbf{C1}) \texttt{FedPop} does not require Bayesian Optimization \cite{zhou2023single} or gradient-based hyperparameter optimization \cite{khodak2021federated}, which saves the communication and computation costs. Besides, \texttt{FedPop} utilizes an \emph{online} evolutionary method (\texttt{Evo}) to update the hyperparameters, i.e., not "training-after-tuning" but "tuning-while-training", which eliminates the need for "retraining" after finding a promising HP-configuration. Note that all procedures in \texttt{FedPop} can be conducted in a parallel and asynchronous manner. (\textbf{C2}) \texttt{FedPop-G} is conducted every $T_g$ communication rounds to mitigate the noise depicted in the validation scores of HP-configurations. Besides, \texttt{FedPop-L} dynamically searches and evaluates the local neighborhood of $\boldsymbol{\beta}$, providing a more informative guidance for the client local HP optimization. Consequently, by enhancing the robustness of $\boldsymbol{\beta}$ to HP perturbation, we aim at improving its robustness against client data Non-IIDness.

\section{Experiments and Analyses}
We conduct an extensive empirical analysis to investigate the proposed method and its viability. Firstly, we compare \texttt{FedPop} with the SOTA and other baseline methods on three common FL benchmarks following \cite{khodak2021federated}. Subsequently, we validate our approach by tuning hyperparameters for complex real-world cross-silo FL settings. Besides, we conduct an ablation study on \texttt{FedPop} to demonstrate the importance of its components. Moreover, we present convergence analysis of \texttt{FedPop} and its promising scalability by training ResNets from scratch on \emph{full-sized} ImageNet-1K with Non-IID label distribution. Finally, we demonstrate the applicability of \texttt{FedPop} when combined with different federated optimization methods.  

\subsection{Benchmark Experiments}
\label{sec:benchmark}
\subsubsection{Datasets Description} 
We conduct experiments on three benchmark datasets on both vision and language tasks: (1) \emph{CIFAR-10} \cite{krizhevsky2009learning}, which is an image classification dataset containing 10 categories of real-world objects. (2) \emph{FEMNIST} \cite{caldas2018leaf}, which includes gray-scale images of hand-written digits and English letters, producing a 62-way classification task. (3) \emph{shakespeare} \cite{caldas2018leaf} is a next-character prediction task and comprises sentences from Shakespeare's Dialogues. 


We investigate 2 different partitions of the datasets: (1) For i.i.d (\emph{IID}) setting, we randomly shuffle the dataset and evenly distribute the data to each client. (2) For non-i.i.d (\emph{NIID}) settings, we follow \cite{khodak2021federated, caldas2018leaf} and assume each client contains data from a specific writer in FEMNIST, or it represents an actor in Shakespeare. For CIFAR-10 dataset, we follow prior arts \cite{zhu2021data, lin2020ensemble} to model Non-IID label distributions using Dirichlet distribution \textbf{$Dir_{x}$}, in which a smaller $x$ indicates higher data heterogeneity. We set the communication budget $(R_t, R_c)$ to $(4000, 800)$ for CIFAR-10 and shakespeare, while $(2000, 200)$ for FEMNIST following \cite{khodak2021federated, caldas2018leaf}. Besides, We adopt 500 clients for CIFAR-10, 3550 clients for FEMNIST, and 1129 clients for Shakespeare. For the coefficients used in \texttt{FedPop}, we set the initial perturbation intensity $\epsilon^0$ to 0.1, the initial resampling probability $p_{re}^0$ to 0.1, and the quantile coefficient $\rho$ to 3. The perturbation interval $T_g$ for \texttt{FedPop-G} is set to $0.1R_c$. Following \cite{khodak2021federated}, we define $\boldsymbol{\alpha} \in \mathbb{R}^3$ and $\boldsymbol{\beta} \in \mathbb{R}^7$, i.e., we tune learning rate, scheduler, and momentum for server-side aggregation (\texttt{Agg}), and learning rate, scheduler, momentum, weight-decay, the number of local epochs, batch-size, and dropout rate for local clients updates (\texttt{Loc}), respectively. More details about the HP search space, dataset descriptions, and model architectures are provided in Appendix.

\subsubsection{Results and Discussion}
In Table \ref{tab:results}, we report the testing accuracy achieved by the final model after performing hyperparameter tuning with different algorithms on three benchmarks. Hereby, we report the results of the \emph{global} model, which is the server model $\boldsymbol{w}$ after the execution of the final communication round, and the \emph{finetuned} model (in the brackets), which is the final global model finetuned on clients local data via \texttt{Loc}($\boldsymbol{\beta}^0, \boldsymbol{w}, T^k$). We observe that \texttt{FedPop}, combined with either \texttt{RS} or \texttt{SHA} as a wrapper, outperforms all the competitors on all benchmarks. For IID settings, the global model tuned on CIFAR-10 with \texttt{FedPop}, with \texttt{RS} or \texttt{SHA} as a wrapper, outperforms the baseline by $2.14\%$ and $2.90\%$, respectively. Likewise, \texttt{FedPop} yields the highest average accuracy on FEMNIST and Shakespeare. For Non-IID settings, \texttt{FedPop} achieves a significant improvement of $3.85\%$ and $4.79\%$ on average compared with \texttt{FedEx} in CIFAR-10, when combined with \texttt{RS} and \texttt{SHA}, respectively. Moreover, we find that the performance improvement of the finetuned model using \texttt{FedPop} surpasses the other baselines. Additionally, we observe that during the tuning procedures, certain trials in the baselines and \texttt{FedEx} fail to converge. We attribute this to their pre-defined and fixed hyperparameter search spaces and values, resulting in higher sensitivity to the hyperparameter initialization that could not be mitigated during the tuning process. This phenomenon is observed via their larger accuracy deviation compared with \texttt{FedPop}, which further highlights the tuning stability of \texttt{FedPop}.

\subsection{Validation on Real-World Cross-Silo FL Systems}
\begin{table}[b]    
    \renewcommand\arraystretch{1.01}
    \vspace{-8pt}
    \caption{Evaluation results on three real-world cross-silo FL benchmarks with feature space distribution shifts. }
    \vspace{-12pt}
    \label{tab:crosssiloresults}
    \setlength{\tabcolsep}{2.5mm}{
    \begin{center}
    \scriptsize
    \centering
    \setlength\tabcolsep{9pt}
    \begin{tabular}{c|ccc}
    
    \toprule
    Tuning Algorithm & PACS  & OfficeHome  & DomainNet \\
    
    \hline
    SHA & \makecell[c]{68.71 {\tiny ±7.38} \\ (76.53 {\tiny ±12.54}) } & \makecell[c]{38.65 {\tiny ±14.82} \\ (57.64 {\tiny ±12.21}) } & \makecell[c]{71.41 {\tiny ±6.56} \\ (79.41 {\tiny ±11.81}) }\\
    \cline{1-4}
    FedEx & \makecell[c]{73.47 {\tiny ±3.06} \\ (80.61 {\tiny ±5.68}) } & \makecell[c]{42.99 {\tiny ±8.72} \\ (58.40 {\tiny ±10.77}) } & \makecell[c]{71.68 {\tiny ±6.13} \\ (78.96 {\tiny ±10.71}) } \\
    \cline{1-4}
    \textbf{FedPop} & \makecell[c]{\textbf{75.17} {\tiny ±1.18} \\ (\textbf{85.37} {\tiny ±2.12}) } & \makecell[c]{\textbf{45.71} {\tiny ±7.64} \\ (\textbf{62.76} {\tiny ±7.38}) } & \makecell[c]{\textbf{73.59} {\tiny ±3.58} \\ (\textbf{81.78} {\tiny ±3.14}) }\\
    
    \bottomrule       
    \end{tabular}
    \vspace{-5pt}
    \end{center}}
\end{table}

As described in Section \ref{sec:relatedwork}, previous hyperparameter tuning algorithms focused on small-scale benchmarks and simple model architectures. To indicate the effectiveness of \texttt{FedPop} on real-world FL applications, we further conduct experiments on three large-scale benchmarks: (1) PACS \cite{li2017deeper}, which includes images that belong to 7 classes from 4 domains Art-Painting, Cartoon, Photo, and Sketch. (2) OfficeHome \cite{venkateswara2017deep}, which contains 65 different real-world objects in 4 styles: Art, Clipart, Product, and Real. (3) DomainNet \cite{peng2019moment}, which is collected under 6 different data sources: Clipart, Infograph, Painting, Quickdraw, Real, and Sketch. All images are reshaped with larger sizes, i.e., 224x224. Following the setting proposed by \cite{li2021fedbn, chen2023fraug}, we apply cross-silo \cite{li2020federated} FL settings and assume each client contains data from one of the sources (domains), but there exist feature distributions shift across different clients (feature space NIID \cite{li2021fedbn}). We use a more complex network architecture, i.e., ResNet-18, as the classification backbone. We set the tuning budget $(R_t, R_c)$ to $(1000, 200)$. More details about the settings are provided in Appendix.

In Table \ref{tab:crosssiloresults}, we report the evaluation results of the target model after tuning by \texttt{SHA} or its combination with \texttt{FedEx} or \texttt{FedPop}. We highlight the performance improvements achieved by the proposed method compared with the competitors, where \texttt{FedPop} surpasses the others up to $2.72\%$ and indicates smaller accuracy deviations. These results indicate the effectiveness of \texttt{FedPop} on real-world FL scenarios with a smaller number of clients, large-scale private datasets, and more complex network architectures. 

\subsection{Ablation Study}
To illustrate the importance of different \texttt{FedPop} components, we conduct an ablation study on CIFAR-10 benchmark considering \emph{IID} and \emph{NIID} settings. The results are shown in Table \ref{tab:ablationresults}. We first notice that applying only one population-based tuning algorithm, i.e., either \texttt{FedPop-L} or \texttt{FedPoP-G}, already leads to distinct performance improvements on the baselines, especially when the client's data are \emph{Non-IID}. Moreover, employing both functions together significantly improves the tuning results, which demonstrates their complementarity.
\begin{table}[t]
    \renewcommand\arraystretch{1.01}
  \centering
     \makeatletter\def\@captype{table}\makeatother\caption{Ablation study for different components in \texttt{FedPop} on CIFAR-10 benchmark.}     
    \vspace{-10pt}
    \label{tab:ablationresults}
    \setlength{\tabcolsep}{1.9mm}{
    \begin{center}
    \scriptsize
    \centering
    \setlength\tabcolsep{10.5pt}
    \begin{tabular}{c|ccc}
    
    \toprule
    \multirow{2}{*}{\makecell[c]{Tuning \\ Algorithm}} & \multicolumn{3}{c}{CIFAR-10} \\
    \cline{2-4}
     ~ & IID & NIID ($Dir_{1.0}$) & NIID ($Dir_{0.5}$) \\
    
    \hline
    RS & \makecell[c]{69.04 {\tiny ±7.38}  } & \makecell[c]{63.47 {\tiny ±3.14}} & \makecell[c]{62.88 {\tiny ±8.13}  } \\
    \cline{1-4}
    FedPop-G & \makecell[c]{70.61 {\tiny ±3.21}  } & \makecell[c]{66.81 {\tiny ±3.24}} & \makecell[c]{65.63 {\tiny ±4.67}} \\
    \cline{1-4}
    FedPop-L & \makecell[c]{70.03 {\tiny ±2.13} }  & \makecell[c]{67.50 {\tiny ±2.06}} & \makecell[c]{64.24 {\tiny ±5.96} }\\
    \cline{1-4}
    FedPop & \makecell[c]{\textbf{71.18} {\tiny ±4.68}} & \makecell[c]{\textbf{68.25} {\tiny ±5.03}  } & \makecell[c]{\textbf{67.01} {\tiny ±4.98} } \\
    
    \bottomrule       
    \end{tabular}
    \vspace{-13pt}
    \end{center}}
\end{table}

\subsection{Analysis on Full-sized NIID ImageNet-1k}
\begin{figure}[b]
\centering
\vspace{-14pt}
  \begin{minipage}[t]{0.45\linewidth}
    \centering
    \includegraphics[width=\linewidth]{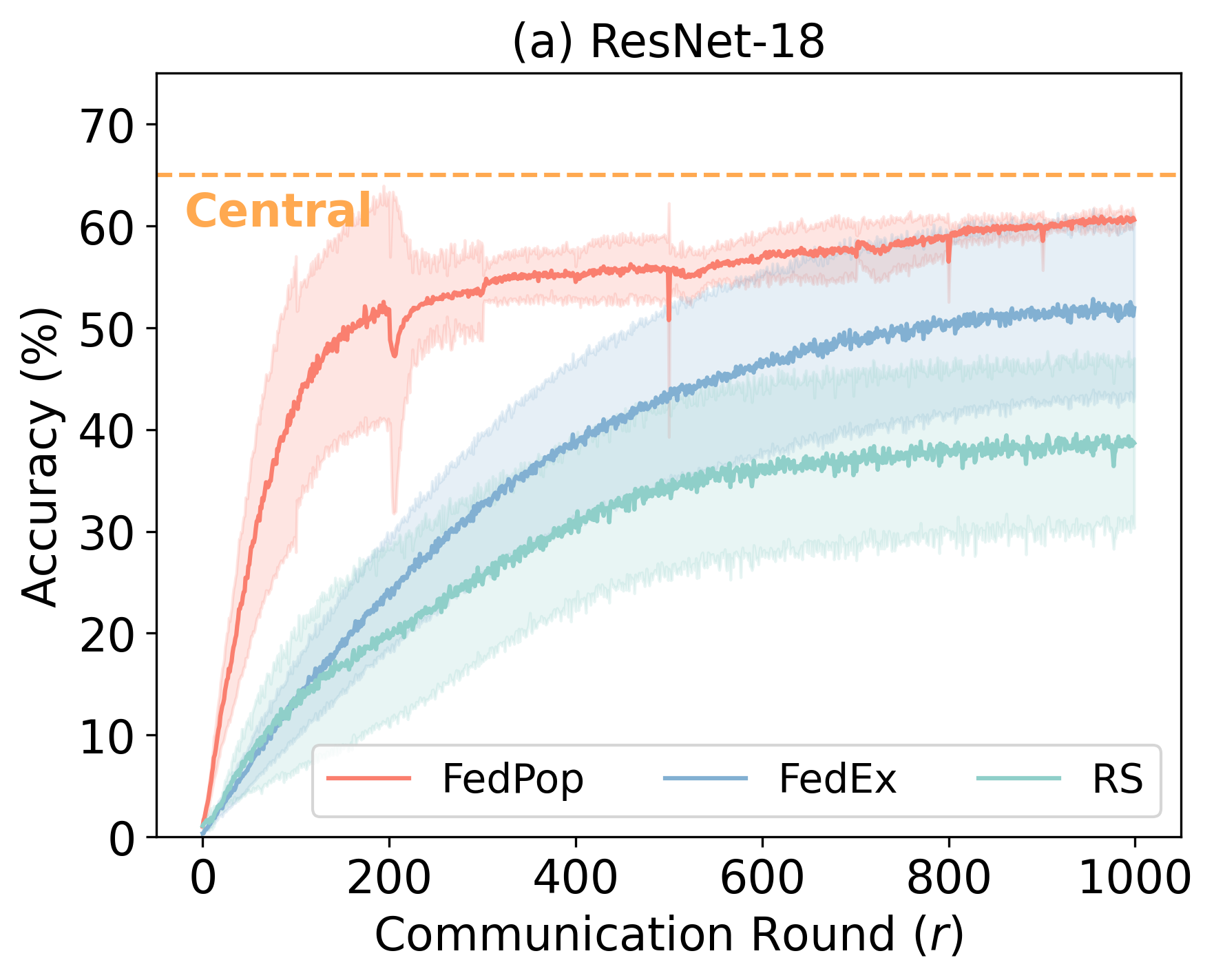} 
  \end{minipage}
  \begin{minipage}[t]{0.45\linewidth}
    \centering
    \includegraphics[width=\linewidth]{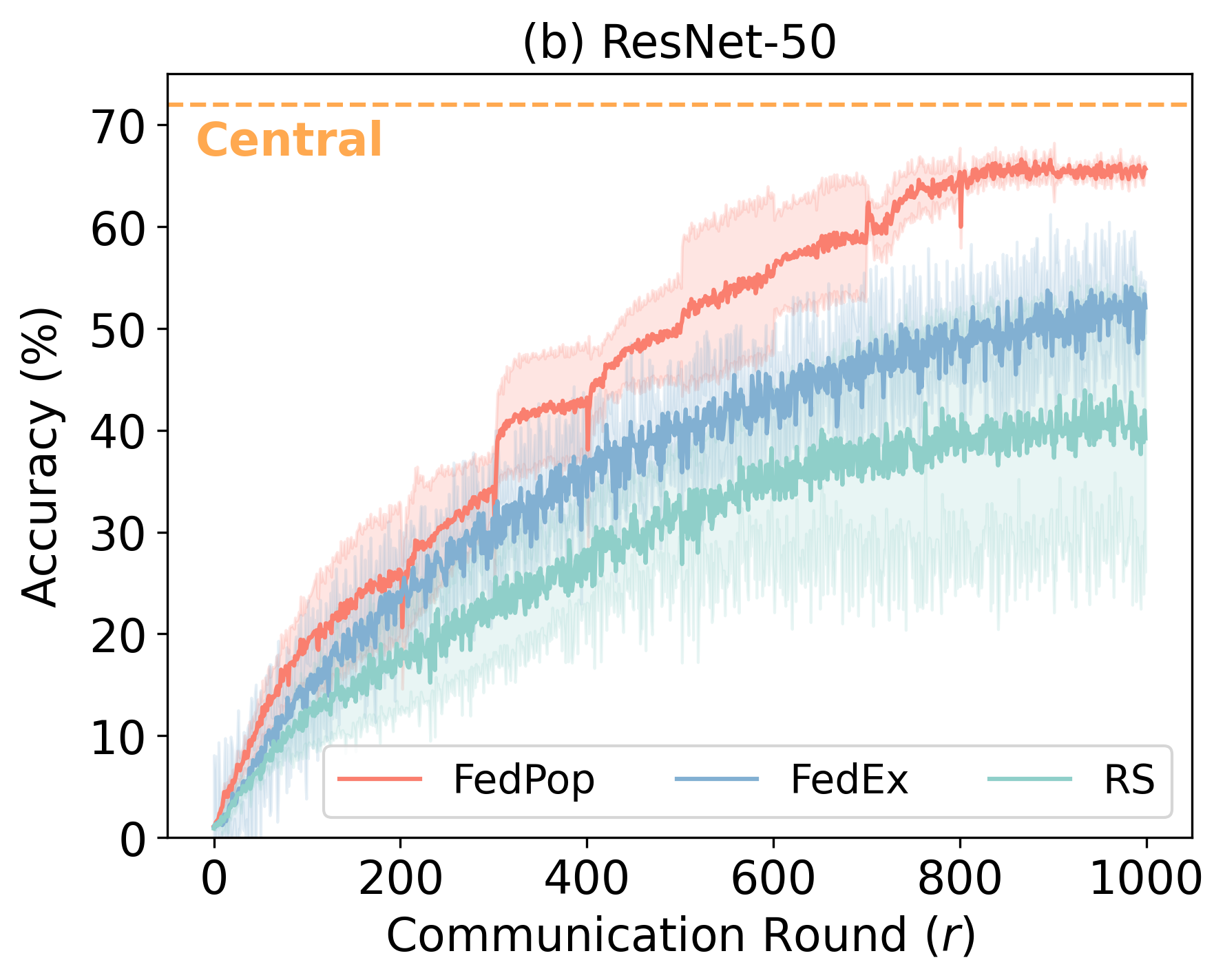} 
  \end{minipage} 
  \vspace{-10pt}
  \captionof{figure}{Convergence analysis of different tuning algorithms on full-sized \emph{Non-IID} ImageNet-1k.}
  \label{fig:converg} 
  \vspace{-8pt}
\end{figure}

To further demonstrate the scalability of \texttt{FedPop}, we display the convergence analysis of \texttt{FedPop} on \emph{full-sized} ImageNet-1K, where we distribute the data among 100 clients with Non-IID label distributions using Dirichlet distribution \textbf{$Dir_{1.0}$}. Hereby, we set $(R_t, R_c) = (5000, 1000)$ and report the average local testing results of the active clients after communication round $r$. We provide more details about the experimental setup in Appendix.

As shown in Figure \ref{fig:converg}, we discover that \texttt{FedPop} already outperforms the others from the initial phase, indicating its promising convergence rate. Besides, we also observe a reduced performance variation in \texttt{FedPop}, which further substantiates the benefits of evolutionary updates in stabilizing the overall tuning procedure. Most importantly, \texttt{FedPop} achieves comparable results with centralized training of the networks, indicating its scalability to tuning HPs for large-scale FL applications.

\subsection{Validation with Different Fed-Opt Methods}
To illustrate the effectiveness of \texttt{FedPop} combined with different Federated Optimization algorithms (\texttt{Fed-Opt}), we further execute experiments by replacing \texttt{FedAvg} with more advanced alternatives: (1) \texttt{FedProx} \cite{li2020federated}, which introduces an proximal term to regularize the client local objective (\texttt{Loc}). The regularization strength is incorporated into $\boldsymbol{\beta}$ as one hyperparameter. (2) \texttt{SCAFFOLD} \cite{karimireddy2020scaffold}, which employs variance reduction to alleviate the weight divergence between the local and global models caused by client drifting. Hereby, the control variates are updated in the server aggregation (\texttt{Agg}), where its learning rate and its scheduler are added in $\boldsymbol{\alpha}$. 

As shown in Table \ref{tab:fedoptablation}, \texttt{FedPop} outperforms the other methods when tuning FL systems with more complex optimization methods, which indicates its promising compatibility and adaptability with various \texttt{Fed-Opt} alternatives.

\begin{table}[t]
  \centering
    \renewcommand\arraystretch{1.01}
    \captionof{table}{Validation of different tuning algorithms using different federated optimization methods.}     
    \vspace{-7pt}
    \label{tab:fedoptablation}
    \scriptsize
    \centering
    \setlength\tabcolsep{5.5pt}
    \begin{tabular}{c|c|ccc}    
    
    \toprule
    \multirow{2}{*}{\makecell[c]{Fed-Opt \\ Method}} &  \multirow{2}{*}{\makecell[c]{HPT \\ Algo.}} & \multicolumn{3}{c}{CIFAR-10} \\
    \cline{3-5}
    ~ & ~ & IID & NIID ($Dir_{1.0}$) & NIID ($Dir_{0.5}$) \\
    
    \hline
    \multirow{3}{*}{FedProx} & RS & \makecell[c]{71.18 {\tiny ±3.34}} & \makecell[c]{66.47 {\tiny ±4.20}} & \makecell[c]{65.04 {\tiny ±2.13}}  \\
    \cline{2-5}
    ~ & FedEx & \makecell[c]{70.82 {\tiny ±4.25}} & \makecell[c]{67.08 {\tiny ±3.18}} & \makecell[c]{65.42 {\tiny ±4.25} } \\
    \cline{2-5}
    ~ & \textbf{FedPop} & \makecell[c]{\textbf{72.97} {\tiny ±3.51}}  & \makecell[c]{\textbf{69.04} {\tiny ±1.68}} & \makecell[c]{\textbf{67.97} {\tiny ±2.56}}\\
    
    \hline
    \hline
    \multirow{3}{*}{SCAFFOLD} & RS & \makecell[c]{70.67 {\tiny ±4.26}} & \makecell[c]{65.98 {\tiny ±4.25}} & \makecell[c]{64.12 {\tiny ±3.62}}  \\
    \cline{2-5}
    ~ & FedEx & \makecell[c]{71.04 {\tiny ±2.15}} & \makecell[c]{66.24 {\tiny ±3.67}} & \makecell[c]{64.98 {\tiny ±3.13}}  \\
    \cline{2-5}
    ~ & \textbf{FedPop} & \makecell[c]{\textbf{71.65} {\tiny ±3.78}}  & \makecell[c]{\textbf{68.88} {\tiny ±3.12}} & \makecell[c]{\textbf{67.31} {\tiny ±3.11}}\\
    \bottomrule           
    \end{tabular}
    \vspace{-12pt}
\end{table}

\section{Conclusion and Outlooks}
In this work, we present a novel population-based algorithm for tuning the hyperparameters used in distributed federated systems. The proposed algorithm \texttt{FedPop} method performs evolutionary updates for the hyperparameters based on the member performance among the population. Its global component \texttt{FedPop-G}, is applicable for tuning hyperparameters used in both server aggregation and client local updates. For a more detailed tuning of hyperparameters specific to client updates, we apply the fine-grained component, \texttt{FedPop-L}. \texttt{FedPop} achieves state-of-the-art results on three common FL benchmarks involving IID or Non-IID data distributions. Moreover, its promising results on real-world FL with feature distribution shifts or with different federated optimization algorithms, as well as on full-sized Non-IID ImageNet-1K, demonstrate its effectiveness and scalability for complex applications. 

\bibliography{aaai25}

\end{document}